%% file: main.tex
\documentclass[10pt,twocolumn,letterpaper]{article}

\usepackage[pagenumbers]{cvpr} 
\usepackage{graphicx} 
\usepackage{caption} 
\usepackage{multirow}
\usepackage{color, colortbl}
\definecolor{Gray1}{gray}{0.9}
\definecolor{Gray2}{gray}{1.0}
\usepackage{booktabs}

\definecolor{TealBlue}{rgb}{1.0, 0.97, 0.91}

\usepackage{pifont}

\usepackage{colortbl} 


\definecolor{cvprblue}{rgb}{0.21,0.49,0.74}

\definecolor{firebrick}{rgb}{0.7, 0.13, 0.13}
\definecolor{darkpastelgreen}{rgb}{0.01, 0.75, 0.24}
\definecolor{deepskyblue}{rgb}{0.0, 0.75, 1.0}
\definecolor{mypink2}{rgb}{.99,.96,.98}
\definecolor{mypink1}{rgb}{.99,.93,.98}
\definecolor{mypink}{rgb}{.99,.90,.98}
\definecolor{mygray}{rgb}{.95,.95,.95}
\definecolor{lv14}{rgb}{0.5,0.5,0.5}

\usepackage[pagebackref,breaklinks,colorlinks,allcolors=cvprblue]{hyperref}
\usepackage[most]{tcolorbox}

\def\paperID{8243} 
\def\confName{CVPR}
\def\confYear{2025}

\title{Document Haystacks:  Vision-Language Reasoning Over Piles of \\1000+ Documents}

\author{
  Jun Chen\textsuperscript{\rm 1}\thanks{Equal contribution}, Dannong Xu\textsuperscript{\rm 2}\footnotemark[1], Junjie Fei\textsuperscript{\rm 1}\footnotemark[1], 
  Chun-Mei Feng\textsuperscript{\rm 3}, Mohamed Elhoseiny\textsuperscript{\rm 1} \\
  \textsuperscript{\rm 1}{King Abdullah University of Science and Technology} \\ 
  \textsuperscript{\rm 2}{The University of Sydney}, 
  \textsuperscript{\rm 3}{IHPC, A*STAR} \\
  \texttt{\{jun.chen,junjie.fei,mohamed.elhoseiny\}@kaust.edu.sa} \\
  \texttt{daxu8019@uni.sydney.edu.au}, \texttt{fengcm.ai@gmail.com}
}

\begin{document}

\twocolumn[{
\maketitle

\renewcommand\twocolumn[1][]{#1}%
\begin{center}
		\includegraphics[width=1.0\linewidth]{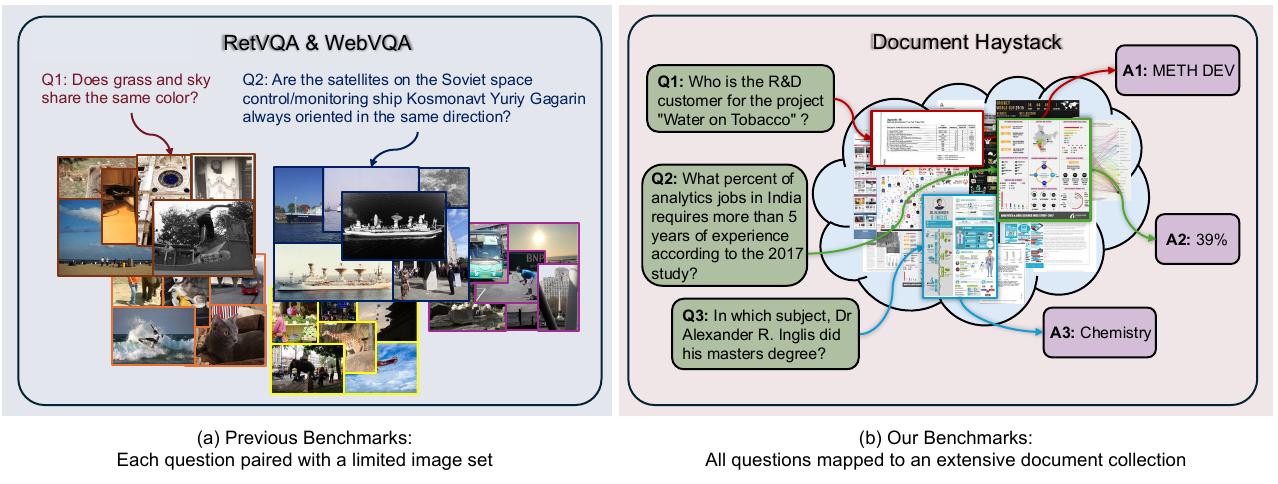}
	\end{center}
    \vspace{-0.5cm}
	\captionof{figure}{\textbf{Comparison between previous and proposed benchmarks.} Given a question as input, all benchmarks aim to retrieve relevant images from an image pool to correctly answer the question. Unlike prior benchmarks like RetVQA~\cite{penamakuri2023retvqa} and WebVQA~\cite{chang2022webqa}, which structure their datasets by pairing each question with a limited set of images (typically $\leq$ 30), our benchmarks, DocHaystack and InfoHaystack, map each question to a substantially larger document collection, scaling up to 1,000 visual documents. This expanded scope more accurately represents large-scale document retrieval scenarios and offers a greater challenge in retrieval accuracy and visual question answering. 
    }
	\label{teaser_figure}
 \vspace{1cm}
}]

\let\thefootnote\relax\footnotetext{$^*$ Equal contribution}
\input{sec/abstract}    
\input{sec/introduction}
\input{sec/related_works}

\input{sec/benchmark}

\input{sec/method}

\input{sec/experiment}

\input{sec/conclusion}

{
    \small
    \bibliographystyle{ieeenat_fullname}
    \bibliography{main}
}


\end{document}

%% file: sec/abstract.tex
\begin{abstract}
Large multimodal models (LMMs) have achieved impressive progress in vision-language understanding, yet they face limitations in real-world applications requiring complex reasoning over a large number of images. Existing benchmarks for multi-image question-answering are limited in scope, each question is paired with only up to 30 images, which does not fully capture the demands of large-scale retrieval tasks encountered in the real-world usages. To reduce these gaps, we introduce two document haystack benchmarks, dubbed DocHaystack and InfoHaystack, designed to evaluate LMM performance on large-scale visual document retrieval and understanding. Additionally, we propose V-RAG, a novel, vision-centric retrieval-augmented generation (RAG) framework that leverages a suite of multimodal vision encoders, each optimized for specific strengths, and a dedicated question-document relevance module. V-RAG sets a new standard, with a 9\% and 11\% improvement in Recall@1 on the challenging DocHaystack-1000 and InfoHaystack-1000 benchmarks, respectively, compared to the previous best baseline models. Additionally, integrating V-RAG with LMMs enables them to efficiently operate across thousands of images, yielding significant improvements on our DocHaystack and InfoHaystack benchmarks. Our code and datasets are available at \url{https://github.com/Vision-CAIR/dochaystacks}

\end{abstract}

%% file: sec/introduction.tex
\section{Introduction}
\label{sec:intro}


Large Multimodal Models (LMMs)~\cite{openai2024gpt4o, google2024gemini, wang2024qwen2, li2024llavaonevision} have made remarkable progress in the vision-language understanding. However, these models still face challenges when tasked with reasoning over extensive collections of images or documents~\cite{visualhaystack}, limiting their effectiveness in real-world applications, such as visual search or querying over large sets of images or documents, like those stored on personal devices or in photo albums. However, there lacks such proper benchmarks to evaluate these capabilities. To address this gap, we introduce the DocHaystack and InfoHaystack benchmarks, designed to evaluate LMMs on large-scale image retrieval and understanding capabilities, pushing the boundaries of LMM performance in complex, real-world scenarios.

The existing multi-image retrieving and reasoning benchmarks are primarily constructed on a small scale, as highlighted in works such as~\cite{penamakuri2023retvqa, talmor2021multimodalqa}. Each question in these benchmarks is paired with only up to 30 images as illustrated in Figure \ref{teaser_figure} (a). However, this limited scope does not align well with real-world scenarios, which often require retrieval and reasoning across hundreds or thousands of images or documents. In contrast, our established benchmarks, depicted in Figure \ref{teaser_figure} (b), allow for querying questions from a large-scale collection of up to 1,000 documents, necessitating that models retrieve and reason over an extensive set of documents for each question. This scale better simulates practical applications and their demands.

The main challenge in constructing such benchmarks is collecting specific questions while ensuring there are no ambiguous answers across a large set of images. Existing datasets, such as those in DocVQA and InfographicVQA~\cite{mathew2021docvqa, mathew2021infographicvqa}, contain numerous ``general" questions, like ``What is the table number?", where answers could be derived from multiple images, leading to non-unique responses. To address this, we implemented a rigorous data filtering pipeline. First, we employed both a large language model (LLM) and human annotators to systematically filter out ``general" questions based on carefully defined criteria. Additionally, we used the LLM to exclude questions relying on generic knowledge, such as ``What is the capital of Missouri?", which can be answered without image context. This approach ensures that the questions in the benchmark can only be answered through specific visual cues from the provided images, maintaining the benchmark's integrity for evaluating image-based understanding.

To enable the current LMMs effectively reason over a large number of images, we propose a vision-centric retrieval-augmented generation (RAG) framework, named V-RAG. V-RAG combines multiple multimodal vision encoders, leveraging each encoder's unique strengths to enhance retrieval accuracy. Additionally, it incorporates an LMM-filter module to assess the relevance of each document to the query, refining the retrieval process by ensuring that only relevant documents are prioritized. This integrated approach allows V-RAG to navigate extensive document collections efficiently. Experimental results demonstrate that V-RAG achieves 9\% and 11\% improvement in Recall@1 on the DocHaystack-1000 and InfoHaystack-1000 compared to previous best text-to-image retrieval methods. Additionally, we found that integrating V-RAG brings GPT-4o over a 55\% acc improvement on DocHaystack-200 and a 34\% acc improvement on InfoHaystack-200, indicating the effectiveness of our V-RAG.

Our contributions are as follows:

\begin{itemize} 
    \item We introduce the Document Haystack benchmarks, including DocHaystack-100/200/1000 and InfoHaystack-100/200/1000, with the most challenging setup consisting of 1,000 documents for each inquiry. These benchmarks advance document retrieval and reasoning tasks by requiring models to navigate and reason across extensive document collections, surpassing prior benchmarks limited to smaller retrieval tasks.
    \item We propose a vision-centric retrieval-augmented generation framework, V-RAG, which enhances the retrieval capabilities of LMMs. V-RAG achieves substantial improvements over previous best text-to-image retrieval methods by 9\% and 11\% on DocHaystack-1000 and InfoHaystack-1000, respectively. 
\end{itemize}





%% file: sec/related_works.tex
\section{Related Works}

\begin{figure*}[t!]
	\begin{center}
        \includegraphics[width=1.0\linewidth]{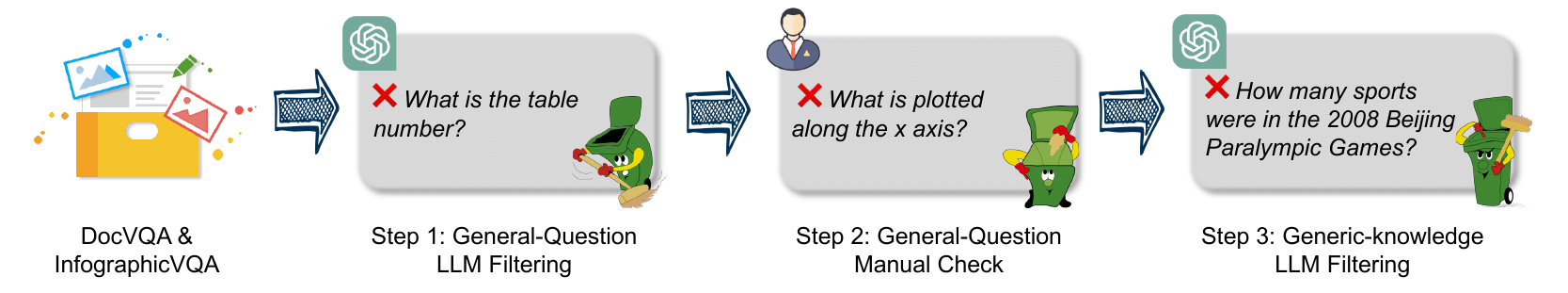}
	\end{center}
	\captionsetup{font=small}
    	\caption{\textbf{Data Curation Pipeline.} Our benchmarks are curated based on the DocVQA and InfographicVQA datasets, following a three-step filtering process to obtain document-specific question-answer pairs. In Step 1, we filter out general questions (e.g., ``What is the table number?"), as these could be answered by multiple documents and lack specificity. Step 2 involves a manual review by human annotators to further remove general questions. In Step 3, we eliminate generic-knowledge questions (e.g., ``How many sports were in the 2008 Beijing Paralympic Games?") that can be answered directly by large language models without requiring image input."}
	\label{data_pipeline}
    \vspace{-0.3cm}
\end{figure*}

\noindent\textbf{VQA benchmarks.} VQA play a critical role in assessing a model’s ability to understand and reason across visual contexts~\cite{balanced_vqa_v2,bordes2024introduction,feng2023vqa4cir}. Traditional VQA datasets typically measure a model's comprehension of object attributes~\cite{hudson2019gqa,krishna2017visual}, spatial relationships~\cite{hudson2019gqa}, as well as its understanding of documents~\cite{mathew2021docvqa,mathew2021infographicvqa}, charts~\cite{masry2022chartqa}, mathematics~\cite{lu2023mathvista,wang2024measuring,zhang2024mathverse}, and open knowledges~\cite{marino2019ok,schwenk2022aokvqa}. Additionally, these benchmarks explore models' knowledge across varied fields, including science and the arts~\cite{lu2022learn,yue2024mmmu}. This broad array of benchmarks has greatly advanced vision-language models by cultivating diverse visual comprehension skills, particularly for modern foundation models in vision-language understanding~\cite{openai2024gpt4v,openai2024gpt4o,google2024gemini, zhu2023minigpt, chen2023minigptv2,lu2022learn, li2024llavaonevision,bai2023qwenvl,wang2023minigemini}. Notably, these benchmarks have primarily focused on question answering within single image or document. In contrast, our benchmark shifts the focus towards retrieval and comprehensive understanding across a large collection of visual documents, presenting new challenges and expanding the scope of visual question answering.

Several previous efforts have tackled the challenge of visual question answering and reasoning across multiple images~\cite{penamakuri2023retvqa, chang2022webqa, bansal2020isvqa, talmor2021multimodalqa, wang2024needle,tanaka2023slidevqa}. For instance, datasets such as MultimodalQA~\cite{talmor2021multimodalqa} and ISVQA~\cite{bansal2020isvqa} require models to have multi-image reasoning abilities. Meanwhile, WebQA~\cite{chang2022webqa} and RetVQA~\cite{penamakuri2023retvqa} involve an additional step where models must first retrieve relevant images from a limited image pool before answering visual questions based on these results. However, these benchmarks are generally constrained to relatively small image pools, where each question is paired with an image set containing up to 30 images. In contrast, our proposed benchmarks, DocHaystack and InfoHaystack, significantly expand this scope by requiring models to retrieve and reason from a much larger set of up to 1,000  documents, presenting a notably greater challenge in retrieval and multi-image reasoning.

\noindent\textbf{Large multimodal models (LMMs).} LMMs have achieved substantial advancements in understanding and reasoning across single or multiple images~\cite{openai2024gpt4o, li2024llavaonevision, google2024gemini, wang2024qwen2, zhu2023minigpt, chen2023minigptv2}. These models have significantly enhanced vision-language understanding across numerous dimensions and applications~\cite{yue2024mmmu, balanced_vqa_v2, zhang2024mathverse, lu2023mathvista}. LMMs benefit primarily from large-scale image-text alignment and extensive language modeling, which emerge them with advanced understanding and reasoning abilities. However, despite these breakthroughs, LMMs still encounter challenges when handling large-scale image or document sets~\cite{visualhaystack}. This difficulty is due to the inherent complexity of processing such complex data. To address this, retrieval-based approaches have been developed to extend the capacity of vision-language models, augmenting their ability to process and reason over a larger number of images.

\noindent \textbf{Retrieval-augmented generation (RAG).} RAG integrates retrieval systems~\cite{clip,zhai2023siglip,openclip,bai2024sentence,feng2023vqa4cir}, with generative models, enhancing them with additional knowledge. While RAG has been extensively explored in language domains~\cite{asai2023selfrag, guu2020realm, luo2023sail, karpukhin2020dense}, its application in vision-language contexts is also advancing. In vision-language RAG, models like MuRAG~\cite{chen2022murag} leverage image-text memory to retrieve top-k neighbors by comparing inner-product similarities. RetVQA~\cite{penamakuri2023retvqa} uses an image-question relevance encoder, combining BERT~\cite{devlin2019bert} and Faster R-CNN~\cite{faster-rcnn} to filter relevant images, while MIRAGE~\cite{visualhaystack} employs a CLIP-based encoder~\cite{clip} to train a retriever. These frameworks extend model capabilities, enabling retrieval and reasoning across hundreds or thousands of images. In contrast, we propose V-RAG, a vision-centric RAG framework that integrates multiple vision encoders to more effectively capture image features, and introduces a LMM-based question-document relevance comparison module. Our results demonstrate that V-RAG surpasses existing methods on our DocHaystack and InfoHaystack benchmarks, setting a new standard for large-scale visual retrieval and reasoning.

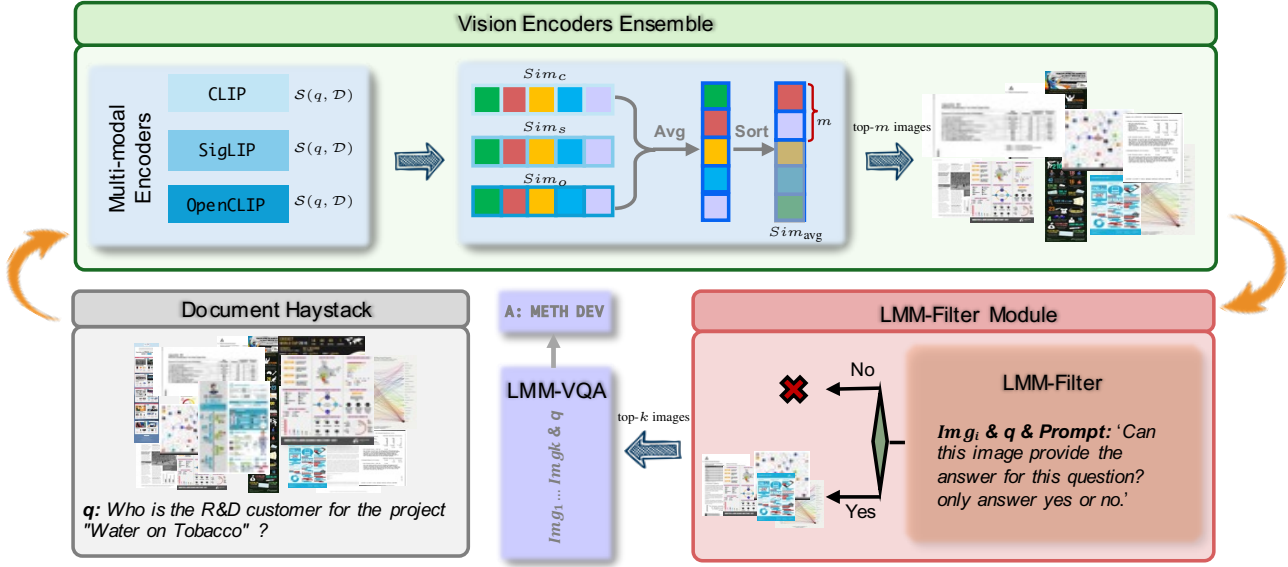
\begin{figure*}[t!]
	\begin{center}
        \includegraphics[width=\linewidth]{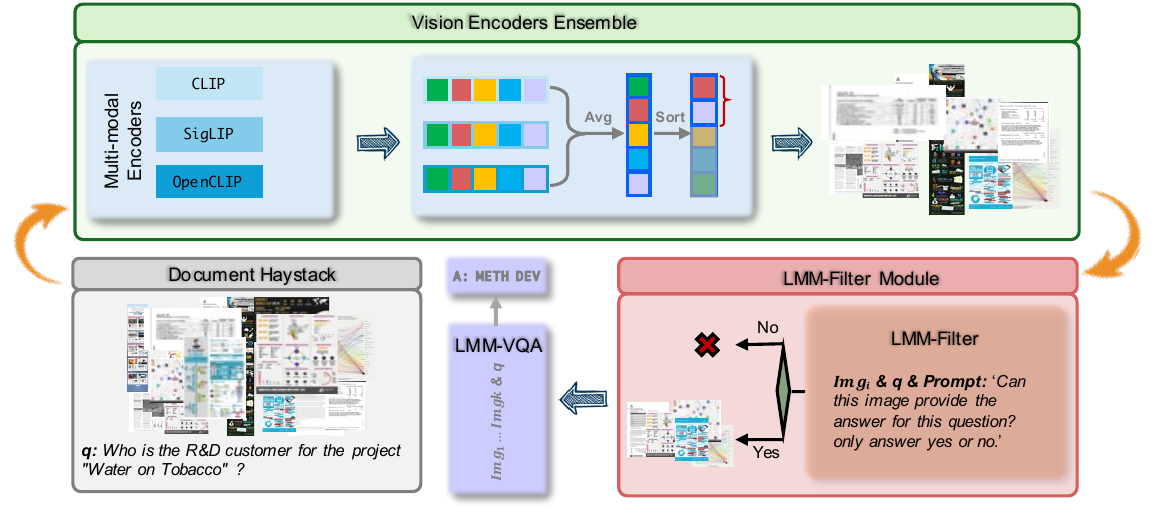}
        \put(-298,170){ \tiny{$Sim_{s}$}}
        \put(-298,190){ \tiny{$Sim_{c}$}}
        \put(-298,150){ \tiny{$Sim_{o}$}}
        \put(-205,130){ \tiny{$Sim_{\text{avg}}$}}
        \put(-187,172){ \tiny{${m}$}}
        \put(-262,60){ \tiny{top-\(k\) images}}
        \put(-173 ,170){ \tiny{top-\(m\) images}}
        \put(-382,182){\tiny{$\mathcal{S}(q, \mathcal{D})$}}
         \put(-382,162){\tiny{$\mathcal{S}(q, \mathcal{D})$}}
          \put(-382,142){\tiny{$\mathcal{S}(q, \mathcal{D})$}}
	\end{center}
	\captionsetup{font=small}
	\caption{\textbf{The V-RAG pipeline workflow.} In the top section, a vision encoder ensemble is used, combining multiple vision models—CLIP, SigLIP, and OpenCLIP—to process a large document haystack. Each encoder computes similarity scores, which are averaged into $Sim_{\text{avg}}$. The top m documents, based on these scores, are selected for further analysis. In the bottom right, the LMM-Filter Module utilizes a pretrained LMM to assess whether each selected document can potentially answer the posed question. This filtering step removes documents that do not match, retaining only relevant ones. Finally, the top k most relevant images are input into the LMM along with the original question $q$ to generate a specific answer.     }
	\label{pipeline}
    \vspace{-0.3cm}
\end{figure*}

%% file: sec/benchmark.tex
\section{DocHaystack and InfoHaystack Benchmarks}
To support effective retrieval and reasoning across extensive document collections, we present two new benchmarks—DocHaystack and InfoHaystack—designed to ensure each question yields a unique, document-specific answer. Derived from DocVQA~\cite{mathew2021docvqa} and InfographicVQA~\cite{mathew2021infographicvqa}, these benchmarks address the challenge of answer ambiguity by selectively curating questions that can only be answered by a single document within a large dataset.

\noindent\textbf{Benchmark construction pipeline.} There exists many general questions in the existing benchmarks and lead to multiple answers for different document context. For example, general questions like ``\texttt{What is the table number?}" may apply to various documents and yield multiple valid answers, while a targeted question like ``\texttt{Who is the reviewer for the article titled} `\texttt{An antithyroid factor in milk}'\texttt{?}" is likely to produce a unique answer, as only a single document or a limited set of documents would contain that information. Therefore, our benchmark construction follows a structured three-step filtering pipeline, illustrated in Figure \ref{data_pipeline}, to ensure high-quality, unique-answer questions. First, we employ a large language model (LLM) to filter out general questions that could generate multiple answers across documents. Next, a manual review step further checks the questions to ensure the data quality. Finally, a generic-knowledge filtering stage refines the dataset further, retaining only questions closely tied to specific document content. 

This carefully designed pipeline, combining LLM-based filtering and human review, effectively curates questions that drive accurate, document-specific retrieval. By focusing on reducing answer ambiguity, DocHaystack and InfoHaystack enhance the precision of retrieval and reasoning in large-scale document processing tasks, providing a valuable tool for the evaluation of retrieval systems. We discuss this data curation pipeline in details as follows:

\noindent\textbf{General-question LLM filtering.} We begin by using the LLM, GPT-4o~\cite{openai2024gpt4o}, to filter out general questions through a set of well-crafted instructions. Leveraging the LLM's strong contextual understanding, this initial filtering step allows us to efficiently process large volumes of data, identifying broad or ambiguous questions that may yield multiple answers across documents. This automated approach significantly enhances the benchmark construction’s efficiency and quality.

To guide the LLM, we first define the task, providing clear distinctions between general and specific questions along with illustrative examples. With this framework, the LLM can then assess each question and determine if it is general or specific. The instructional format is as follows:LLM i

\textit{You are an evaluator tasked with identifying if a question is specific or general. A general question seeks commonly known or widely applicable information without unique identifiers, e.g., ``Who is the person standing in the ground?" A specific question, however, requests unique information about a particular individual, event, or object, e.g., ``What is the Social Security Number of Charles Yarbrough?" Based on these definitions, determine if the following question is general or specific: \{question\}.}

\noindent \textbf{General-question manual review.} After the initial LLM filtering, we conduct a manual review of the questions that were classified as specific. This manual process involves two key steps to ensure answer uniqueness and benchmark quality.

In the first step, we examine each question to confirm it contains unique identifiers—such as names, dates, titles, or other specific attributes—suggesting a document-specific answer. This careful check helps identify questions with clear, unique markers that direct the retrieval process to a single document.

In the second step, we verify the uniqueness of each answer to eliminate any remaining ambiguity. Although specific identifiers are present, questions may still be prone to ambiguity, such as with common names or recurring book titles. To address this, we employ a refined verification process. First, we use Optical Character Recognition (OCR)~\cite{tesseract} to extract all text from images in the dataset. We then search for occurrences of the unique identifiers retained from the first step across other documents. If matches are found, a manual review is conducted to ensure no alternative valid answers exist. This comprehensive approach minimizes the possibility of a single question mapping to multiple answers, enhancing the precision and reliability of our benchmarks.


\noindent \textbf{Generic-knowledge filtering.} In DocVQA and InfographicVQA tasks, certain questions—such as ``\texttt{How many sports were in the 2008 Beijing Paralympic Games?}"—can be answered based on general knowledge accessible to a large language model, without relying on the image content. This introduces a language bias when using LMMs for visual question answering, as it shifts the focus away from image-based reasoning. To address this, we filter out these general-knowledge questions, ensuring that evaluation emphasizes vision-based understanding and that models rely primarily on visual content to generate accurate answers.

To implement this, we developed an LLM-based evaluation pipeline that detects and excludes such questions. For each question, we prompt an LLM with ``\texttt{\{question\}, answer briefly.}". After receiving a response, we compare it to the ground-truth answer using another LLM. If the response matches the ground truth, we classify the question as general knowledge-related and remove it, thereby isolating questions that truly require visual document understanding. As shown in Table \ref{knowledge_filter}, GPT-4o accurately answers 26.4\% of DocVQA questions and 
54.9\% of InfographicVQA questions directly, a rate significantly higher than that of open-source LLMs. Therefore, we select GPT-4o to filter out the questions that can be directly answered by the GPT-4o model. Overall, this process is to ensure that the evaluation reflects the necessity of vision-based comprehension.


\begin{table}[t!]
\centering
\resizebox{1.0 \linewidth}{!}{
    \begin{tabular}{c ccc}
    \toprule
     & GPT-4o & LLaVA-OneVision& Qwen2-VL\\
    \cmidrule(r){1-1} 
\cmidrule(lr){2-2} \cmidrule(lr){3-3} \cmidrule(lr){4-4}
    DocVQA & 26.4\% & 4.7\% & 3.4\% \\
    InfographicVQA & 54.9\% & 13.4\% & 11.3\% \\
    \bottomrule
    \end{tabular}
}
\caption{\textbf{Percentage of questions answerable by LMMs without vision input.} We evaluate GPT-4o, LLaVA-Onevision, and Qwen2-VL on their ability to answer questions directly from our dataset without requiring vision input. The reported percentage reflects the proportion of examples that can be answered solely through language understanding.}
\label{knowledge_filter}
\end{table}

\noindent\textbf{Final dataset profile.} After a rigorous three-stage data filtering process, we retained 109 questions from DocVQA and 155 questions from InfographicVQA, associated with 59 and 66 documents that provide the evidence, respectively. To assess retrieval performance at scale, we introduce two benchmarks: DocHaystack-1000 and InfoHaystack-1000, where each question requires retrieving relevant content from a set of 1,000 documents. Given the challenge this scale presents to current LMMs, particularly in terms of context length limitations, we also construct two smaller benchmarks: DocHaystack-100/200 and InfoHaystack-100/200. These benchmarks allow direct input of all associated images into the context, enabling evaluation of models' long-context comprehension ability. For training set, we also construct a dataset comprising 2,835 questions similarly, with 899 from DocVQA and 1,936 from InfographicVQA, to support robust learning and generalization for the multi-image reasoning tasks.

\begin{figure}[t!]
	\begin{center}
        \includegraphics[width=1.0\linewidth]{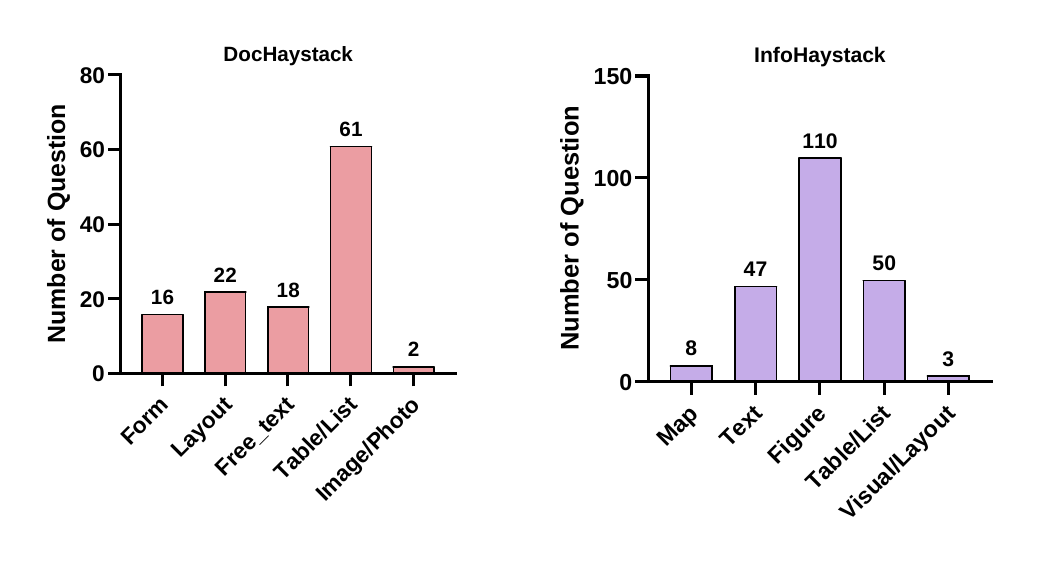}
        \vspace{-1.0cm}
	\end{center}
	\captionsetup{font=small}
	\caption{\textbf{Question type analysis.} We analyze the distribution of question types of DocHaystack and InfoHaystack. Each benchmark categorizes the data into 5 different types. }
	\label{question_type}
     \vspace{-0.4cm}
\end{figure}

\noindent\textbf{Question type analysis.} The types of questions represent the types of the evidence required for accurate answers. In Figure \ref{question_type}, we illustrate the distribution of question types across our dataset to provide insights into its structure. Following the classification system used in DocVQA and InfographicVQA, we categorize questions accordingly (note that a single question may fall into multiple categories). As shown in the figure, the DocHaystack benchmark places a greater emphasis on Table/List and Layout understanding, whereas InfoHaystack primarily targets Figure, Text, and Table comprehension.

%% file: sec/method.tex
\section{Method}

Current large multimodal models (LMMs) face substantial challenges when reasoning across hundreds or thousands of images, due not only to context length limitations but also to the inherent complexity of the task. This issue is particularly pronounced in our benchmarks, which contain 1k document files requiring high-resolution images to capture and interpret small-font text effectively. To enable LMMs to perform reasoning over a substantial number of documents, we introduce a vision-centric retrieval-augmented generation (V-RAG) framework. V-RAG efficiently retrieves a reduced  set of relevant documents, allowing the LMM to focus on a manageable subset for deeper understanding, as illustrated in Figure \ref{pipeline}. In the following section, we provide a detailed description of the V-RAG pipeline.


\noindent \textbf{Task definition.} Given a question $q$ and a collection of \( N \) documents \( \mathcal{D} = \{D_1, \dots, D_N\} \), the V-RAG framework aims to retrieve the top-\( k \) most relevant documents to support LMMs understanding and answering the question $q$. V-RAG accomplishes this through a two-step retrieval process designed to effectively identify and rank relevant documents for each question.

\noindent \textbf{Vision encoder ensemble.} Document files often contain a mix of text, symbols, and visual elements across various scales, requiring vision encoders to capture a comprehensive understanding of these complex structures. To efficiently handle this diversity, we represent each document as an image and utilize an ensemble of vision encoders, including CLIP~\cite{clip}, SigLIP~\cite{zhai2023siglip}, and OpenCLIP~\cite{openclip}, each bringing distinct strengths to the image understanding, as depicted in Figure \ref{pipeline}. For example, the ConvNext encoder~\cite{liu2022convnet} from OpenCLIP~\cite{openclip} is particularly effective for high-resolution image encoding. We compute the similarity score between each question \( q \) and all documents in the document set \( \mathcal{D} \) according to Equation \ref{similarity}, with similarity scores from each encoder represented as \( Sim_{c} \), \( Sim_{o} \), and \( Sim_{s} \) respectively.

\begin{equation} \mathcal{S}(q, \mathcal{D}) = { cos(\phi_{\text{t}}(q), \phi_{\text{v}}(D_j)) \mid D_j \in \mathcal{D} }, \label{similarity} \end{equation}
where $\mathcal{S}$ denotes the computing the similarity between the query $q$ and a collection of documents $\mathcal{D}$. $cos$ denotes the cosine similarity. $\phi_\text{t}$ denotes the text encoder, and $\phi_\text{v}$ denotes the vision encoder.

To derive a final relevance score, we calculate the average similarity \(Sim_{\text{avg}}\) for each question-image pair by combining \(Sim_{c}\), \(Sim_{o}\), and \(Sim_{s}\). We then rank the images based on \(Sim_{\text{avg}}\) in descending order, selecting the top-\(m\) most relevant images according to their similarity scores.


\input{table/retrieval}

\noindent \textbf{LMM-filter module.} To refine the selection of top-\(m\) relevant images further, we introduce a LMM-based question-image relevance assessment module. This module evaluates the relevance between each question and the top-\(m\)  images identified in the first filtering step. Specifically, we pair each image with the question text and input them into an open-source vision-language model, prompting, ``Can this image provide answers to this question? Respond only with yes or no''. We only retain the question-image pairs that are identified as  "yes" from LMM, and remove other irrelevant images.

\noindent \textbf{LMM-VQA module.} Achieving high top-1 ranking accuracy in image retrieval is challenging, so we retain the top-\(k\) images from the LMM-filtered ranking list and present them to the LMM-VQA to improve the likelihood of including relevant images. We input these top-\(k\) images alongside the question into the LMM-VQA (see Figure \ref{pipeline}), which then generates the answer directly. To enhance robustness against visual distractors, the LMM-VQA can be further optimized, as analyzed in the experiment section.


%% file: table/retrieval.tex
\begin{table*}[t!]
\centering
\setlength\tabcolsep{12pt}
\scalebox{0.75}{
\begin{tabular}{lccccccccc}
\toprule
 & \multicolumn{3}{c}{\textbf{DocHaystack-100}}&  \multicolumn{3}{c}{\textbf{DocHaystack-200}} & \multicolumn{3}{c}{\textbf{DocHaystack-1000}} \\
 & R@1 & R@3 & R@5   & R@1 & R@3& R@5 &R@1 & R@3& R@5  \\
\cmidrule(r){1-1} 
\cmidrule(lr){2-4} \cmidrule(lr){5-7} \cmidrule(lr){8-10}
BM25 (OCR) & 63.30 & 75.23 & 79.82 & 65.14 & 71.56 & 75.23 & 56.88& 66.06& 69.72 \\
Jina-CLIP~\cite{koukounas2024jina} & 16.51 & 31.19 & 41.28 & 9.17 & 24.77 & 30.28 & 3.67 & 7.34 & 12.84 \\
Nomic-Embed-Vision~\cite{nussbaum2024nomic} & 16.51 & 24.77 & 28.44 & 13.76 & 21.10 & 25.69 & 1.83 & 2.75 & 6.42 \\
CLIP~\cite{clip} & 46.79 & 65.14 & 69.72 &  44.04 & 58.72 &65.14 &23.85& 41.28& 45.87 \\
SigLIP~\cite{zhai2023siglip} & 51.38 & 67.89 & 76.15 &47.71 & 63.30 & 70.64 & 33.03& 49.54& 57.80\\
OpenCLIP~\cite{openclip} & 58.72 & 75.23 & 79.82 & 56.88 & 70.64 & 75.23 & 34.86& 49.54 & 57.80 \\

\rowcolor{TealBlue}
\textbf{V-RAG (ours)} & \textbf{81.65} &  \textbf{88.99} & \textbf{88.99}   &  \textbf{77.98} & \textbf{84.40}  & \textbf{84.40}  & \textbf{66.06} & \textbf{77.98}  &  \textbf{78.90} \\
\midrule
 & \multicolumn{3}{c}{\textbf{InfoHaystack-100}} & \multicolumn{3}{c}{ \textbf{InfoHaystack-200}} &
 \multicolumn{3}{c}{\textbf{InfoHaystack-1000}} \\
  & R@1 & R@3 & R@5 &  R@1 & R@3& R@5 & R@1 & R@3& R@5\\
\cmidrule(r){1-1} 
\cmidrule(lr){2-4} \cmidrule(lr){5-7} \cmidrule(lr){8-10}

BM25 (OCR) & 56.77 & 65.81 & 70.97 & 51.61 & 65.16 & 69.03 & 38.71 & 51.61 & 58.06 \\
Jina-CLIP & 43.23 & 51.61 & 58.06 & 36.77 & 46.45 & 51.61 & 23.87 & 33.55 & 37.42 \\
Nomic-Embed-Vision & 34.84 & 50.32 & 56.77 & 30.97 & 43.23 & 48.39 & 20.65 & 30.97 & 35.48 \\
CLIP & 69.68 & 78.71 & 85.81 & 65.16& 77.42 & 81.94 &45.81& 64.52& 70.32\\
SigLIP & 58.06 & 71.61 & 80.00& 55.48 & 67.74 & 76.77 &39.35 & 55.48& 61.94 \\
OpenCLIP & 72.26 & 85.16 & \textbf{92.90}&  66.45 & 81.94 & \textbf{89.03} & 53.55 & 65.81 & 72.90  \\
\rowcolor{TealBlue}
\textbf{V-RAG (ours)}  & \textbf{79.35} & \textbf{90.97} & \textbf{92.90} & \textbf{74.84} & \textbf{88.39}  &  88.39 & \textbf{64.52}& \textbf{74.19} &  \textbf{78.06} \\
\bottomrule
\end{tabular}
}
\caption{ \textbf{Retrieval Results.} We compare our V-RAG model with other text-to-image and text-to-text (using OCR) retrieval methods across both benchmarks. V-RAG consistently outperforms baseline models on Recall@1, Recall@3, and Recall@5 metrics. Notably, V-RAG leverages an ensemble of text-to-image models along with a large multimodal model in a two-stage filtering approach. Top-performing values in each column are highlighted in \textbf{bold}.}
\label{tab:retrival}
\vspace{-0.3cm}
\end{table*}

%% file: sec/experiment.tex
\section{Experiments}
\label{sec:formatting}

In the experiments section, we will primarily describe our training setup, covering evaluation metrics, baseline models, and the fine-tuning procedure for the LMM-VQA model. We also present the main experimental results along with an ablation study to provide further insights.

\subsection{Training setup}

\paragraph{Metric.}
In our evaluation of the DocHaystack and InfoHaystack benchmarks, we employ a model-based assessment by leveraging GPT-4o-mini~\cite{openai2024gpt4o} to accurately determine whether the model predictions match target answers. This method uses a carefully structured prompt to facilitate GPT-4o-mini's evaluation of answer correctness. We empirically found that the model-based evaluation achieves higher consistency and alignment with human judgment. Additional details on the prompt design are provided in the Appendix. 

For the document retrieval evaluation, we report the baseline results using Recall@1, Recall@3, and Recall@5 metrics. These metrics enable a thorough assessment of retrieval accuracy across varying levels of precision.

\input{table/baseline}
\noindent \textbf{Baselines.} In our experiment, we have evaluated several open and closed-sourced vision-language models on the retrieval and VQA performance. For the large multimodal model, we used the \textit{gpt-4o-2024-08-06} version of GPT-4o~\cite{openai2024gpt4o}, the \textit{LLaVA-OneVision-Qwen2-7b-OV-HF} version of LLaVA-OneVision~\cite{li2024llavaonevision}, and the \textit{Qwen2-VL-7B-Instruct} version of Qwen2-VL~\cite{bai2023qwenvl}. For computing the text-to-image similarities, we employed the \textit{Jina-CLIP-v1}~\cite{koukounas2024jina} variant, \textit{Nomic-Embed-Vision-v1.5}~\cite{nussbaum2024nomic} variant, CLIP~\cite{clip} ViT-L/14@336 variant, for SigLIP~\cite{zhai2023siglip}, the ViT-SO400M/14@384 variant, and for OpenCLIP~\cite{openclip}, the ConvNeXt-XXL@1024 variant as well as text-based method, BM25.

In our V-RAG setting, we apply \textit{LLaVA-OneVision-Qwen2-7b-OV-HF} for the LMM-filter module and \textit{Qwen2-VL-7B-Instruct} for the LMM-VQA module. We select $m$ as 60 and $k$ as 5 in our experiment.

\noindent \textbf{Optimizing the LMM-VQA module.} To improve the robustness of the LMM-VQA model in handling visual question answering with multiple distractor images, we further fine-tune the model using our curated training data.

During this fine-tuning process, we introduce 1–10 randomly sampled distractor images for each question, creating a challenging setting that encourages the model to focus on relevant content amid a mix of positive and negative images. The fine-tuning is conducted with a batch size of 32 and a peak learning rate of 1e-4 over a single epoch. Additionally, we leverage LoRA~\cite{hu2021lora} with a rank of 8 to efficiently adapt the model’s parameters during training.

\subsection{Main Experimental Results}
We evaluated a range of open-source and closed-source vision-language models for VQA tasks. We also evaluate several text-to-image  and text-to-text (with OCR) retrieval models to evaluate their retrieval capabilities on our benchmarks. More detailed performance analysis are described in the following sections.

\noindent \textbf{Retrieving results.} The retrieval results in Table~\ref{tab:retrival} demonstrate the superiority of our proposed V-RAG framework over several baseline methods across both DocHaystack and InfoHaystack benchmarks. V-RAG consistently achieves the highest Recall@1, Recall@3, and Recall@5 scores on most categories, indicating its robust retrieval capabilities. Notably, V-RAG outperforms  text-based retrieving models such as BM25 and also the text-to-image retrieval models like jina-clip, CLIP, SigLIP, and OpenCLIP by substantial margins, especially on the DocHaystack-100 subset, where it reaches Recall@1 of 81.65\% and Recall@5 of 88.99\%. This pattern continues for larger datasets (DocHaystack-1000), where V-RAG remains competitive, achieving Recall@1 of 66.06\%. It achieves the top performance across all recall metrics on DocHaystack. For InfoHaystack benchmarks, V-RAG also outperforms other models, particularly on InfoHaystack-100 and InfoHaystack-200, where it receives Recall@1 of 74.84\% and 64.52\%, higher than previous best by 8\% and 11\%, respectively. This consistent performance advantage highlights the effectiveness of V-RAG’s ensemble of multiple vision encoders, allowing it to capture more granular details and improve retrieval accuracy over large multimodal models.

\begin{figure*}[t!]
	\begin{center}
        \includegraphics[width=1.0\linewidth]{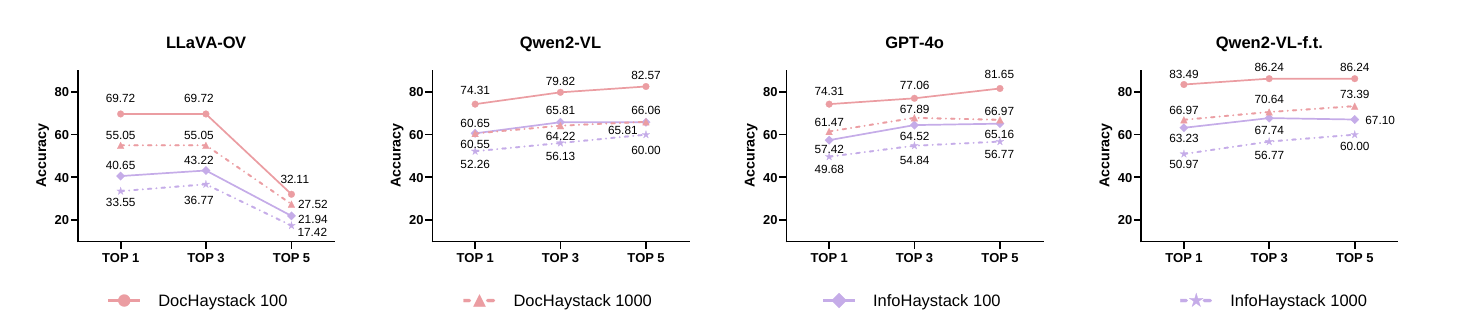}
        \vspace{-1.0cm}
	\end{center}
	\captionsetup{font=small}
	\caption{\textbf{Top-k selection ablation analysis for LMM-VQA.} We demonstrate the results for LLaVA, Qwen2-VL, GPT-4o and also the finetuned Qwen2-VL model on the DocHaystack-100/1000 and InfoHaystack-100/1000 benchmarks. All the models are integrated with our V-RAG framework. We show the VQA accuracy performance for each ablation. }
	\label{retrieval_ablation}
\end{figure*}


\input{table/retrieval_ablation}

\noindent \textbf{Visual question answering (VQA) results.} The table presents VQA results for the DocHaystack and InfoHaystack benchmarks across varying dataset sizes (100, 200, 1000) using different multimodal models, both independently and in combination with the V-RAG framework. The results show that Qwen2-VL fine-tuned with V-RAG (Qwen2-VL-f.t.+V-RAG) achieves the highest scores across most benchmarks, with particularly notable performance on DocHaystack-100 (86.24) and InfoHaystack-100 (67.10), indicating superior retrieval and VQA capabilities in these scenarios. When V-RAG is added to other models, substantial improvements are observed, demonstrating the framework’s efficacy in enhancing retrieval accuracy. For instance, GPT-4o's performance increases significantly with V-RAG, particularly for DocHaystack-100 and -200. The analysis highlights that V-RAG integration generally boosts performance across models, with Qwen2-VL-f.t.+V-RAG standing out as the top performer on both benchmarks, especially for the larger 1000-document tasks where retrieval accuracy is more challenging. This suggests that V-RAG’s vision-centric, retrieval-augmented approach is highly effective for large-scale multimodal document understanding.

The table also shows that the DocHaystack-1000 and InfoHaystack-1000 present significant challenges for current LMMs. The drop in performance for larger document sets, with top accuracy only reaching 73.39\% for DocHaystack-1000 and 60.00\% for InfoHaystack-1000, underscores the difficulty our benchmarks.


\subsection{Ablation Studies}

\noindent \textbf{Ablation study on Top-k Selection.} This figure presents the top-k selection ablation analysis for LMM-VQA across four models: LLaVA-OV, Qwen2-VL, GPT-4o, and the fine-tuned Qwen2-VL (Qwen2-VL-f.t.), evaluated on the DocHaystack-100/1000 and InfoHaystack-100/1000 benchmarks. The analysis reports VQA accuracy as a function of top-k selection (Top 1, Top 3, and Top 5). Overall, accuracy tends to improve with larger k-values, suggesting that offering more retrieval options positively impacts model performance. However, for LLaVA-OV, there is a marked decrease in performance at top-5, indicating that this model struggles to process multiple images at this scale.




\noindent \textbf{Ablation study on the V-RAG framework components.} The ablation study in Table~\ref{tab:retrieval_ablation} highlights the contributions of each component in the V-RAG framework on the DocHaystack-1000 and InfoHaystack-1000 benchmarks. Using CLIP alone yields low performance (e.g., Recall@1 of 23.85\% on DocHaystack-1000 and 45.81\% on InfoHaystack-1000), indicating its limited retrieval capability on its own. Adding SigLIP and OpenCLIP incrementally improves results.

The highest performance is achieved when all three encoders are combined with the VLM-filter module, leading to Recall@1 scores of 66.06\% on DocHaystack-1000 and 64.52\% on InfoHaystack-1000. This setup also achieves the top Recall@1, Recall@3 and Recall@5 values, demonstrating that the VLM-filter is essential for refining the ensemble outputs and significantly improving retrieval accuracy. These results confirm that each module contributes to V-RAG’s overall effectiveness.




%% file: table/baseline.tex
\begin{table}
\centering
\resizebox{ \linewidth}{!}{
    \begin{tabular}{l ccc ccc}
    \toprule
    \multirow{2}{*}{\textbf{Model}} & \multicolumn{3}{c}{\textbf{DocHaystack}} & \multicolumn{3}{c}{\textbf{InfoHaystack}}  \\
    ~ & 100 &  200 & 1000 & 100 & 200 & 1000 \\    

\cmidrule(r){1-1} 
\cmidrule(lr){2-4} \cmidrule(lr){5-7}
    LLaVA-OV~\cite{li2024llavaonevision} & -    & -    & -    & -    & -    & -    \\
    GPT-4o~\cite{openai2024gpt4o}        & 27.52 & 23.85 & -    & 23.87 & 20.00 & -    \\
    Gemini~\cite{google2024gemini}       & 50.46 & 48.62 & -    & 29.03 & 21.94 & -    \\
    Qwen2-VL~\cite{wang2024qwen2}        & 41.28 & 12.84 & -    & 20.00 & 14.19 & -    \\
    MIRAGE~\cite{visualhaystack}         & 3.67  & 3.67  & 2.75     & 7.74  & 7.10 & 6.45     \\
    \midrule
    LLaVA-OV+V-RAG & 69.72 & 65.14 & 55.05 & 43.22 & 41.94 & 36.77 \\
    GPT-4o+V-RAG   & 81.65 & 72.48 & 66.97 & 65.16 & 63.23 & 56.77 \\
    Gemini+V-RAG   & 73.39  &  65.14 & 58.72 & 57.42 & 57.42  & 47.10 \\
    \rowcolor{Gray2}
    Qwen2-VL+V-RAG & 82.57 & 74.31 & 66.06 & 65.81 & 65.81 & \textbf{60.00} \\
    \midrule
    \rowcolor{TealBlue}
    Qwen2-VL-f.t.+V-RAG & \textbf{86.24} & \textbf{79.82} & \textbf{73.39} & \textbf{67.10}& \textbf{67.74} & \textbf{60.00}\\
    \bottomrule
    \end{tabular}
}
\caption{\textbf{The VQA results for the DocHaystack and InfoHaystack.} We evaluate with many closed-source and open-source multimodal model, and also integrating them with our V-RAG retrieval framework. - denotes that those models can not be inferred due to their token context constraints. To enable GPT-4o and Qwen2-VL to process hundreds of images, we employ low-resolution mode and adjust image size for compatibility.}
\label{tab:maintable}
\vspace{-0.6cm}
\end{table}

%% file: table/retrieval_ablation.tex
\begin{table*}[t!]
\centering
\setlength\tabcolsep{10pt}

\scalebox{0.7}{
\begin{tabular}{cccccccccc}
\toprule
 \multirow{2}{*}{\textbf{CLIP}}  & \multirow{2}{*}{ \textbf{SigLIP}} &\multirow{2}{*}{\textbf{OpenCLIP}} & \multirow{2}{*}{\textbf{VLM-filter}}& \multicolumn{3}{c}{\textbf{DocHaystack-1000}} & \multicolumn{3}{c}{\textbf{InfoHaystack-1000}} \\
 &&&& R@1 & R@3 & R@5    & R@1 & R@3 & R@5  \\

\cmidrule(r){1-1} 
\cmidrule(lr){2-2} \cmidrule(lr){3-3} \cmidrule(lr){4-4} \cmidrule(lr){5-7} \cmidrule(lr){8-10} 
 

{\color{green} \checkmark} & {\color{red} \ding{55}} & {\color{red} \ding{55}} & {\color{red} \ding{55}}& 23.85 & 41.28 & 45.87 & 45.81 & 64.52 & 70.32 \\

{\color{red} \ding{55}} & {\color{green} \checkmark} & {\color{red} \ding{55}} & {\color{red} \ding{55}}& 33.03 & 49.54 & 57.80 & 39.35 & 55.48 & 61.94 \\

{\color{red} \ding{55}}& {\color{red} \ding{55}}& {\color{green} \checkmark} & {\color{red} \ding{55}} &  34.86 &49.54 & 57.80&53.55 & 65.81&72.90 \\

{\color{green} \checkmark} & {\color{green} \checkmark} & {\color{red} \ding{55}}& {\color{red} \ding{55}} & 40.37 & 59.63 & 62.39 & 59.35 & 67.74 &74.19 \\

{\color{green} \checkmark} & {\color{green} \checkmark} & {\color{green} \checkmark} & {\color{red} \ding{55}} & 
42.20 & 66.06 & 77.48 & 56.13 & 70.97 & \textbf{78.06}  \\
{\color{green} \checkmark} & {\color{green} \checkmark} & {\color{green} \checkmark} & {\color{green} \checkmark} & {\cellcolor{TealBlue}\textbf{66.06}} &{\cellcolor{TealBlue}\textbf{77.98}} &{\cellcolor{TealBlue}\textbf{78.90}} & {\cellcolor{TealBlue}\textbf{64.52}} & {\cellcolor{TealBlue}\textbf{74.19} }& {\cellcolor{TealBlue}\textbf{78.06} } \\

\bottomrule
\end{tabular}
}
\caption{\textbf{Ablation study on the V-RAG framework components.} We quantify the impact of each module for the Recall@1, Recall@3 and Recall@5 retrieval performance on the DocHaystack-1000 and InfoHaystack-1000 for our V-RAG framework.}
\vspace{-0.1cm}
\label{tab:retrieval_ablation}
\end{table*}

%% file: sec/conclusion.tex
\section{Conclusion}

In this work, we introduced the DocHaystack and InfoHaystack benchmarks to evaluate LMMs for retrieving and reasoning across large-scale documents. Our benchmarks providing a more rigorous and realistic assessment of large multimodal models in real-world, large-scale retrieval scenarios. To tackle these challenges, we proposed V-RAG, a vision-centric retrieval-augmented generation framework that significantly enhances retrieval precision and overall VQA performance. V-RAG achieves this through an ensemble of vision encoders and a specialized relevance filtering module, enabling improved accuracy across diverse visual inputs. Experimental results indicate that integrating V-RAG enables both open-source and closed-source LMMs to achieve superior performance in large-scale image retrieval and complex reasoning tasks.
